\documentclass[conference]{IEEEtran}
\IEEEoverridecommandlockouts
\usepackage{cite}
\usepackage{amsmath,amssymb,amsfonts}
\usepackage{algorithmic}
\usepackage{graphicx}
\usepackage{textcomp}
\usepackage{xcolor}
\def\BibTeX{{\rm B\kern-.05em{\sc i\kern-.025em b}\kern-.08em
    T\kern-.1667em\lower.7ex\hbox{E}\kern-.125emX}}
    
\usepackage{multirow}
\usepackage{url}

\newcommand{\msq}{\ensuremath{\mathrm{m/s^{2}}}}
\newenvironment{compactItemize}{\begin{list}{$\bullet$}
			{\setlength{\topsep}{1mm}\setlength{\itemsep}{0.25mm}
				\setlength{\parsep}{0.1mm}
				\setlength{\itemindent}{0mm}\setlength{\partopsep}{0mm}
				\setlength{\labelwidth}{15mm}
				\setlength{\leftmargin}{4mm}}}{\end{list}}

\newcommand{\etal}{\textit{et al.}}

\usepackage[font={small}]{caption}
\usepackage{fancyhdr}

\begin{document}

\fancypagestyle{firststyle}
{
   \fancyhf{}
   \fancyhead[C]{This is the authors' copy of the paper to appear in the Proceedings of the 3rd International Workshop on Foundation Models for Cyber-Physical Systems \& Internet of Things (FMSys'26).}
   \renewcommand{\headrulewidth}{0pt} 
}

\title{Agentic Driving Coach: Robustness and Determinism of Agentic AI-Powered Human-in-the-Loop Cyber-Physical Systems}

\author{\IEEEauthorblockN{Deeksha Prahlad}
\IEEEauthorblockA{
\textit{Arizona State University}\\
Tempe, AZ, United States \\
dprahlad@asu.edu}
\and
\IEEEauthorblockN{Daniel Fan}
\IEEEauthorblockA{
\textit{Arizona State University}\\
Tempe, AZ, United States \\
danielfa@asu.edu}
\and
\IEEEauthorblockN{Hokeun Kim}
\IEEEauthorblockA{
\textit{Arizona State University}\\
Tempe, AZ, United States \\
hokeun@asu.edu}
}
\IEEEaftertitletext{\vspace{-2.5\baselineskip}}

\maketitle

\thispagestyle{firststyle}

\begin{abstract}
Foundation models, including large language models (LLMs), are increasingly used for human-in-the-loop (HITL) cyber-physical systems (CPS) because foundation model-based AI agents can potentially interact with both the physical environments and human users.
However, the unpredictable behavior of human users and AI agents, in addition to the dynamically changing physical environments, leads to uncontrollable nondeterminism.
To address this urgent challenge of enabling agentic AI-powered HITL CPS, we propose a reactor-model-of-computation (MoC)-based approach, realized by the open-source Lingua Franca (LF) framework. 
We also carry out a concrete case study using the agentic driving coach as an application of HITL CPS.
By evaluating the LF-based agentic HITL CPS, we identify practical challenges in reintroducing determinism into such agentic HITL CPS and present pathways to address them.
\end{abstract}


\section{Introduction}
Foundation models~\cite{bommasani2021opportunities}, general-purpose large-scale machine learning (ML) models such as large language models (LLMs), have been rapidly adopted by various computing systems.
Foundation models are a critical component for agentic AI~\cite{fang2025comprehensive} systems, autonomous systems with goal-directed reasoning. 
Foundation models and agentic AI have great potential to benefit cyber-physical systems (CPS), such as for complex reasoning~\cite{han2025toward} or analyzing human activity~\cite{xu2025exploring}.
This is particularly true for the human-in-the-loop (HITL) CPS~\cite{schirner2013future}, as shown in recent use cases of foundation models in HITL CPS for autonomy, personalization, optimization, safety, human-like interaction, actuation on behalf of humans, etc.~\cite{fernandes2025people,banerjee2024cps,yang2024llm}

However, using foundation models and LLMs for CPS also raises risks and safety concerns~\cite{xu2024llm}.
There have been recent efforts to ensure the robustness and safety of LLM-enabled CPS via formal methods~\cite{hafez2025safe} or fine-tuning~\cite{kimura2024efficiency}.
Nevertheless, the important problem of ensuring \textbf{determinism} has seldom been explored in the usage of foundation models and agentic AI in CPS, especially in HITL CPS.
\textbf{Determinism}, broadly understood as the property of having a single outcome for a given initial state and inputs, plays a key role in enhancing the safety of CPS~\cite{lee2021determinism}, enabling composability, repeatable testing, predictable behavior, and fault detection.

To address this gap, in this paper, we explore pathways toward ensuring determinism in agentic AI-powered HITL CPS leveraging the \textbf{reactor model of computation (MoC)}~\cite{lohstroh2019reactors} via an open-source framework \textbf{Lingua Franca (LF)}~\cite{lohstroh2021toward} for deterministic, concurrent runtime coordination of the reactor MoC.
Specifically, we build a model of an example HITL CPS in LF, \textbf{agentic driving coach}, together with the models of a human driver, a car, and a physical environment in LF.
Our agentic driving coach demonstrates how we can reconcile the determinism requirement in HITL CPS with the foundation model and human with intrinsic nondeterminism by leveraging reactor MoC and its constructs.

\section{Background and Research Challenges}

Our proposed HITL CPS model builds on \textbf{Lingua Franca (LF)}~\cite{lohstroh2021toward} as a model-based design framework for the reactor MoC.
The \textbf{reactor MoC}, a deterministic variant of the actor MoC~\cite{actor}, is composed of reactors~\cite{lohstroh2019reactors} for modeling concurrent components~\cite{lohstroh2019actors}.
The reactor MoC achieves determinism via communication through ports, deterministic ordering and scheduling of reactions that implement application-specific logic, and the hierarchical structure allowing a complete topological view of sub-reactors within each composite reactor.
The reactor MoC also provides mechanisms (a) to handle deadline violations as exceptions for repeatable and analyzable fault behaviors~\cite{LeeAkellaPaladinoBando2025Deadlines} and (b) to model various latencies in CPS as delayed events with constructs called logical delays for explicit tradeoffs between consistency and availability~\cite{lee2023consistency}.

A growing body of research focuses on leveraging the capabilities of integrating CPS and advanced machine intelligence~\cite{cpsai, CPSgames, responsibilityCPS}.
HITL CPS with agents has been studied for applications across agriculture, robotics, and autonomous systems~\cite{wang2023voyager, rivadeneira2024unified, sreeram2021human, gil2019designing}. 
Specifically, LLMs and agentic AI offer strong capabilities for contextual reasoning to generate adaptive behavioral guidance in dynamic CPS environments~\cite{burgueno2024human}.
However, ensuring determinism in HITL CPS is challenging due to asynchronous events and interaction with the physical world~\cite{lee2021determinism,lohstroh2024deterministic}. 
Srinivasan \etal~\cite{srinivasan2025dura} propose DURA-CPS, which investigates multi-agent evaluation of LLM behavior in simulated CPS. Prior work has demonstrated the use of LF in CPS domains, including automotive systems and AUTOSAR-based safety-critical event coordination~\cite{LFRisk, LFAUTOSAR}. But to the best of our knowledge, no prior work has clearly addressed determinism in HITL CPS integrated with agentic AI.

Many ML- and AI-based techniques are used to build a driving coach. AutoCoach~\cite{autocoach} presents an ML-based driving coach that models driver behavior using supervised learning with support vector machines. 
Fu \etal~\cite{fu2024drive} utilize LLMs for driving-based scenarios and leverage LLMs' decision-making capabilities. 
A recent assessment of AI tools for driver monitoring~\cite{AISCE} highlights the integration of driver-facing sensors with roadway data to predict SCE. However, these approaches~\cite{autocoach, fu2024drive} do not consider determinism as a primary challenge in such systems.

\begin{figure}
    \centering
    \includegraphics[width=0.6\columnwidth]{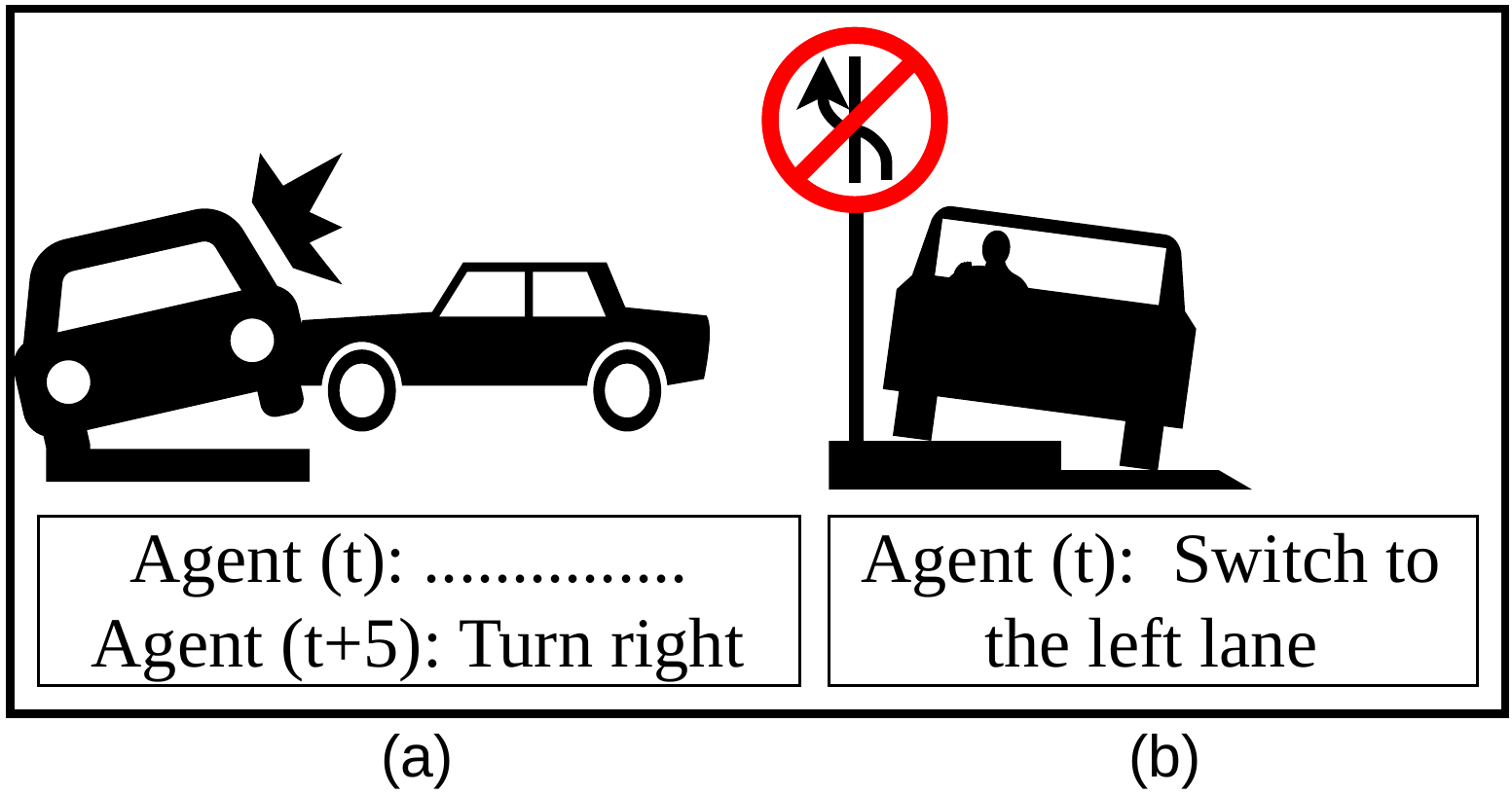}
    \caption{Challenges faced by an agentic driving coach as an agentic AI-powered HITL CPS. (a) The agent delivers an instruction to the driver too late (at time $t+5$). (b) The agent's instruction is incorrect, although delivered timely at time $t$.}
    \label{fig:challenges}
    \vspace{-10pt}
\end{figure}

\figurename~\ref{fig:challenges} demonstrates \textbf{illustrative challenges} introduced by nondeterminism in timing and quality of LLM inferences using an agentic driving coach as an example of agentic HITL CPS. 
\figurename~\ref{fig:challenges}a describes a scenario where a correct (turn-right instruction by the agentic coach), but a delayed response (e.g., 5 seconds after passing the intersection) can lead to a traffic accident.
On the contrary, a timely but incorrect instruction (e.g., switch to the left lane instead of the right lane), which can be due to hallucination in LLMs, can also trigger unsafe behavior of the driver, as shown in \figurename~\ref{fig:challenges}b.


These challenges of response time (\figurename~\ref{fig:challenges}a) and accuracy or correctness (\figurename~\ref{fig:challenges}b) are intrinsic to LLMs and agentic AI, and their resolution is essential to enable agentic AI-powered HITL CPS.
Variable latency in LLM inferences arises from architectural design, model parameters, numerical precision, and quantization.
Also, LLM models with larger parameters and higher precision produce quality responses that are critical for HITL CPS.
Thus, our primary research goal is to surface \textbf{this intrinsic nondeterminism} in CPS modeling and design, analyze and embrace it, and make the overall system model deterministic despite the inherent nondeterminism in agentic HITL CPS, rather than eliminate it completely.

\section{Proposed Approach}

\begin{figure}
    \centering
    \includegraphics[width=0.65\columnwidth]{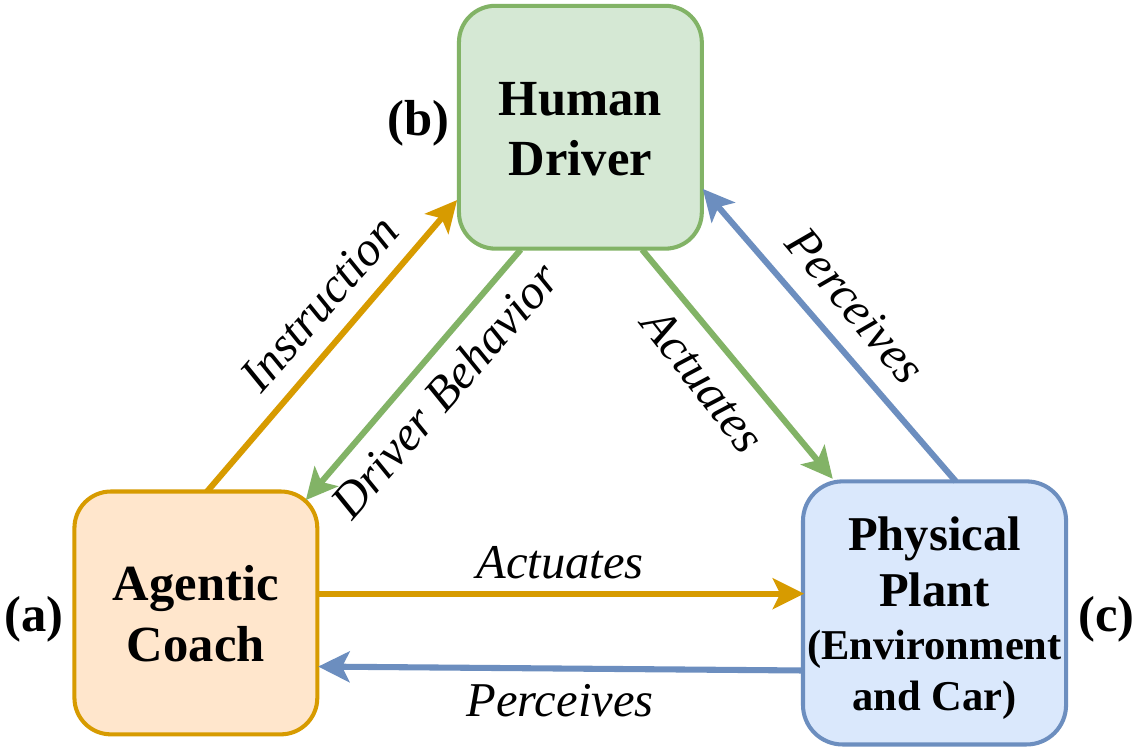}
    \caption{HITL CPS depicting the interaction between the three key components, (a) agentic coach, (b) human driver, and (c) physical plant that includes the environment and car.}
    \label{fig:pa}
    \vspace{-10pt}
\end{figure}

Our proposed approach aims to address the challenge of achieving deterministic modeling of an agentic driving coach as a HITL CPS, given its inherent nondeterminism.
\figurename~\ref{fig:pa} represents our high-level interaction model between the key components of our target system, the agentic driving coach. 
\textbf{Agentic coach} (\figurename~\ref{fig:pa}a) perceives the \textbf{human driver} (\figurename~\ref{fig:pa}b)'s behavior, as well as the physical world (\figurename~\ref{fig:pa}c), including the \textbf{environment}, and the \textbf{car}.
Based on this perception, the \textbf{coach} instructs the \textbf{driver} and sets the actuation command for the \textbf{car}.
The \textbf{driver} receives the instruction from the \textbf{coach} while perceiving the \textbf{environment} on their own.
Based on the \textbf{driver}'s intent and instruction by the \textbf{coach}, the \textbf{driver} actuates (e.g., accelerates, brakes, steers) the \textbf{car}.

\subsection{Problem Formulation}


Based on the key interactions within the agentic HITL CPS, as shown in \figurename~\ref{fig:pa}, we model the system over $n$ discrete time steps, where time $t \in \{1,\dots,n,\dots\}$. Let $x_i$ denote the initial state of the system. At each time step $t$, the system receives three inputs: driver behavior $i_h(t)$, environmental and car input $i_c(t)$, and the agentic coach input $i_a(t)$. The resulting system behavior is defined as:
\begin{equation}
    y(t) = F(x_i, i_h(t), i_c(t), i_a(t))
    \label{eq:system}
\end{equation}
\noindent
where $y(t)$ represents the resulting system behavior at time $t$ and $F(\cdot)$ denotes the system function.
Equation~\eqref{eq:system} states that, for any given initial state and sequential order of driver and coach input, there exists exactly one resulting system behavior.

Note that we treat the \textbf{intrinsic determinism of the HITL CPS as a series of inputs}, and model the system such that it behaves deterministically for each set of inputs.


\begin{figure*}
    \centering
    \includegraphics[width=0.95\textwidth]{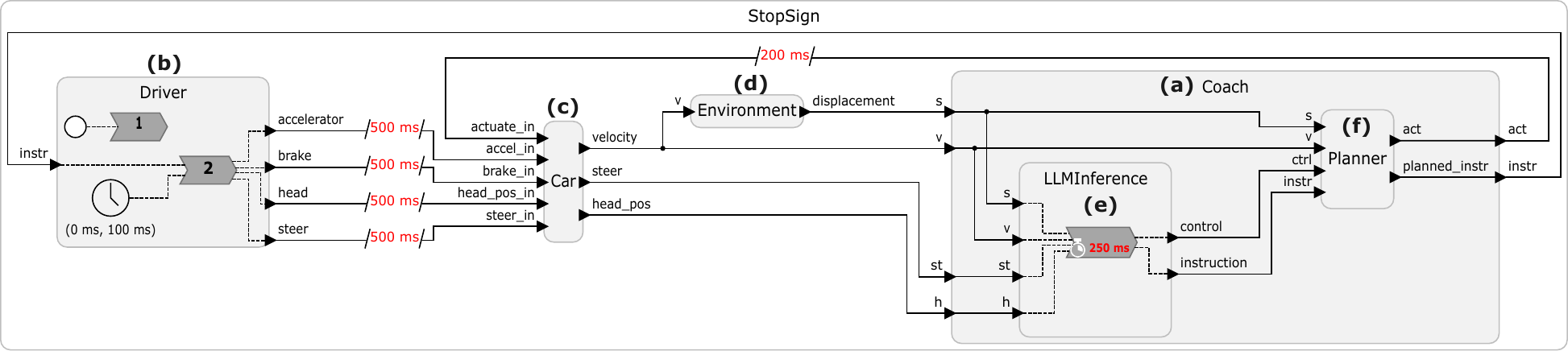}
    \vspace{-5pt}
    \caption{Our proposed reactor model implemented using Lingua Franca, targeting the stop sign scenario, consists of four key reactors: (a) \texttt{Coach},  (b) \texttt{Driver}, (c) \texttt{Car}, and (d) \texttt{Environment}. \texttt{Coach} includes two sub-reactors: (e) \texttt{LLMInference} and (f) \texttt{Planner}.}
    \label{fig:reactormodel}
    \vspace{-10pt}
\end{figure*}
\subsection{Reactor model}
\label{sec:reactor}

Based on our formalization in Equation~\eqref{eq:system}, we model the agentic HITL CPS shown in \figurename~\ref{fig:pa}, with the \textbf{reactor MoC} using the LF framework to ensure deterministic behavior.
\figurename~\ref{fig:reactormodel} describes our reactor MoC consisting of four reactors, namely \textbf{\texttt{Coach}} (corresponding to the agentic coach in \figurename~\ref{fig:pa}a), \textbf{\texttt{Driver}} (for the human driver in \figurename~\ref{fig:pa}b), as well as \textbf{\texttt{Environment}} and \textbf{\texttt{Car}} (for the physical plant in \figurename~\ref{fig:pa}c).
These reactors include reactions (represented by the chevrons in \figurename~\ref{fig:reactormodel}) for implementing the application-specific logic in the target programming language, Python. 

The \textbf{\texttt{Driver}} reactor models the human driver's behavior such that
upon receiving instructions (via an \textit{input port}, \texttt{instr}) from the \textbf{\texttt{Coach}} reactor, \textbf{\texttt{Driver}} reacts based on environmental perception and produces control \textit{outputs} such as \texttt{accelerator}, \texttt{brake}, \texttt{head} position/orientation, and \texttt{steer} actions via \textit{output ports}.
We model continuous human perception using a discrete-event timer (a clock-shaped icon within \textbf{\texttt{Driver}}) that triggers every 100 milliseconds (ms).
We also model the time delay between human perception and physical response~\cite{han2021driver} using \textit{logical delays} of 500ms (marked red on connections between \textbf{\texttt{Driver}} and \textbf{\texttt{Car}} in \figurename~\ref{fig:reactormodel}).
\begin{table}[]
\centering
\footnotesize
\caption{Abstracted driver behaviors used in our agentic driving coach model. $a$ represents the acceleration of the car.}
\label{tab:driver}
\vspace{-5pt}
\begin{tabular}{|c|c|c|}
\hline
\textbf{Driver behavior}& \textbf{Behavior Types}&\textbf{Details}\\
\hline
\multirow{4}{*}{\textbf{Accelerator}}
 & Coasting & $a = -0.1$ \msq   \\
 & Cruise & $a = 0.1$ \msq   \\
 & Normal Acceleration & $a = 2$ \msq   \\
 & Strong Acceleration & $a = 4$ \msq   \\
\hline

\multirow{2}{*}{\textbf{Brake}}
 & Gentle braking & $a = -3.0$ \msq   \\
 & Emergency braking & $a = -9.0$ \msq  \\
\hline

\textbf{Head position}  & Left / Right / Center& \\
\hline

\textbf{Steering angle} & Left / Right / Center &\\
\hline

\end{tabular}
\end{table}

\textbf{\texttt{Car}} receives accelerator, brake, and steering/head positions from \textbf{\texttt{Driver}}, as well as the actuation commands from \textbf{\texttt{Coach}}, via \textit{input ports}.
Then, \textbf{\texttt{Car}} produces the resulting car's physical states over \textit{output ports}, including velocity, steering, and driver's head position (e.g., using the car's interior camera).
Our approach adopts a discrete abstraction of driver pedal behavior commonly used in car driving models and control studies \cite{gillespie2021fundamentals}.
We abstract the car's states based on the driver's behaviors (e.g., how much the car accelerates) as shown in the \tablename~\ref{tab:driver}.
We calculate the car's velocity for each $n$ discrete time step using the following Equation~\eqref{eq:velocity}:
\begin{equation}
    v_{(n+1)} = v_{(n)} + a \times \Delta t
    \label{eq:velocity}
\end{equation}
\noindent
where the notations, $v_n$, $v_{n+1}$, $a$, and $\Delta t$, are defined in \tablename~\ref{tab:notation}.

The \textbf{\texttt{Environment}} reactor models and outputs the car's driving environment as displacement (or distance) driven by the car,
based on this velocity input, as defined in Equation~\eqref{eq:displ}.
\begin{equation}
    s_{(n+1)} = s_{(n)} + v_{(n)} \times \Delta t
    \label{eq:displ}
\end{equation}
\noindent
where again, the notations are defined in \tablename~\ref{tab:notation}.
\begin{table}[]
\centering
\footnotesize
\caption{Notations used in the Equations~\eqref{eq:velocity} and~\eqref{eq:displ}.}
\label{tab:notation}
\vspace{-5pt}
\begin{tabular}{|c|l|}
\hline
\textbf{Notation} & \textbf{Description} \\
\hline
$v_n$ & Velocity of the car at time $n$ \\
\hline
$v_{n+1}$ & Velocity of the car at time $n+1$ \\
\hline
$s_n$ & Displacement of the car at time $n$ \\
\hline
$s_{n+1}$ & Displacement of the car at time $n+1$ \\
\hline
$a$ & Net acceleration of the car \\
\hline
$\Delta t$ & Time interval between two consecutive time steps \\
\hline
\end{tabular}
    \vspace{-10pt}
\end{table}

\begin{table}[]
\centering
\footnotesize
\caption{Control signals generated by the LLM for mode transitions in the \textbf{\texttt{Planner}} reactor.}
\label{tab:ctrl_tokens}
\vspace{-5pt}
\begin{tabular}{|c|c|}
\hline
\textbf{Control Signal} & \textbf{Description} \\ \hline

\textbf{NONE} & No intervention \&  driver \\ &behavior remains within velocity bounds. \\ \hline

\textbf{WARNING} & Driver behavior deviates \\ &from velocity bounds. \\ \hline

\textbf{ACTUATE} & Safety limits are violated \& \\& the system triggers actuation. \\ \hline

\end{tabular}
\end{table}

The \textbf{\texttt{Coach}} reactor is a composite reactor that models the agentic driving coach, consisting of two sub-reactors, \textbf{\texttt{LLMInference}} and \textbf{\texttt{Planner}}.
\textbf{\texttt{Coach}} perceives the driver’s behavior as well as the physical world via \textit{input ports}. 
Then, \textbf{\texttt{Coach}} internally triggers LLM inference modeled by the \textbf{\texttt{LLMInference}} reactor, which generates a control signal and an instruction, moderated by \textbf{\texttt{Planner}}.

\textbf{\texttt{LLMInference}}'s reaction (logic implemented in Python, denoted as a chevron) calls the LLM that is prompted by the structured prompt to be explained later in Section~\ref{sec:prompt}.
This reaction uses the \textbf{deadline-handling mechanism} of the reactor MoC and LF to detect excessive and unsafe delays as deadline violations.
The language constructs for detecting deadline violation are twofold:
\begin{compactItemize}
\item the \textbf{deadline handler} (like an exception handler) definition,
denoted by the stopwatch icon and a default deadline value of 250ms (marked red) to be set based on the worst-case inference time measured in Section~\ref{sec:deadline_selection}, and
\item the use of LF's \texttt{lf\_check\_deadline()} API function~\cite{LeeAkellaPaladinoBando2025Deadlines} to check if the deadline is violated and if violated, call the deadline handler.
\end{compactItemize}
The deadline handler implements fallback mechanisms to ensure safety (e.g., emergency braking and pulling over), a common approach in autonomous vehicles (e.g., Waymo)~\cite{schwall2020waymo}.

The reaction in \textbf{\texttt{LLMInference}} invokes the text generation of up to 30 tokens with a control signal (\texttt{control}) as described in \tablename~\ref{tab:ctrl_tokens} (either \textit{NONE}, \textit{WARNING}, or \textit{ACTUATE}), followed by a sequence of tokens for a driving (\texttt{instruction}), to be fed into \texttt{Planner}.

\begin{figure}[b]
    \centering
    \vspace{-5pt}
    \includegraphics[width=0.65\columnwidth]{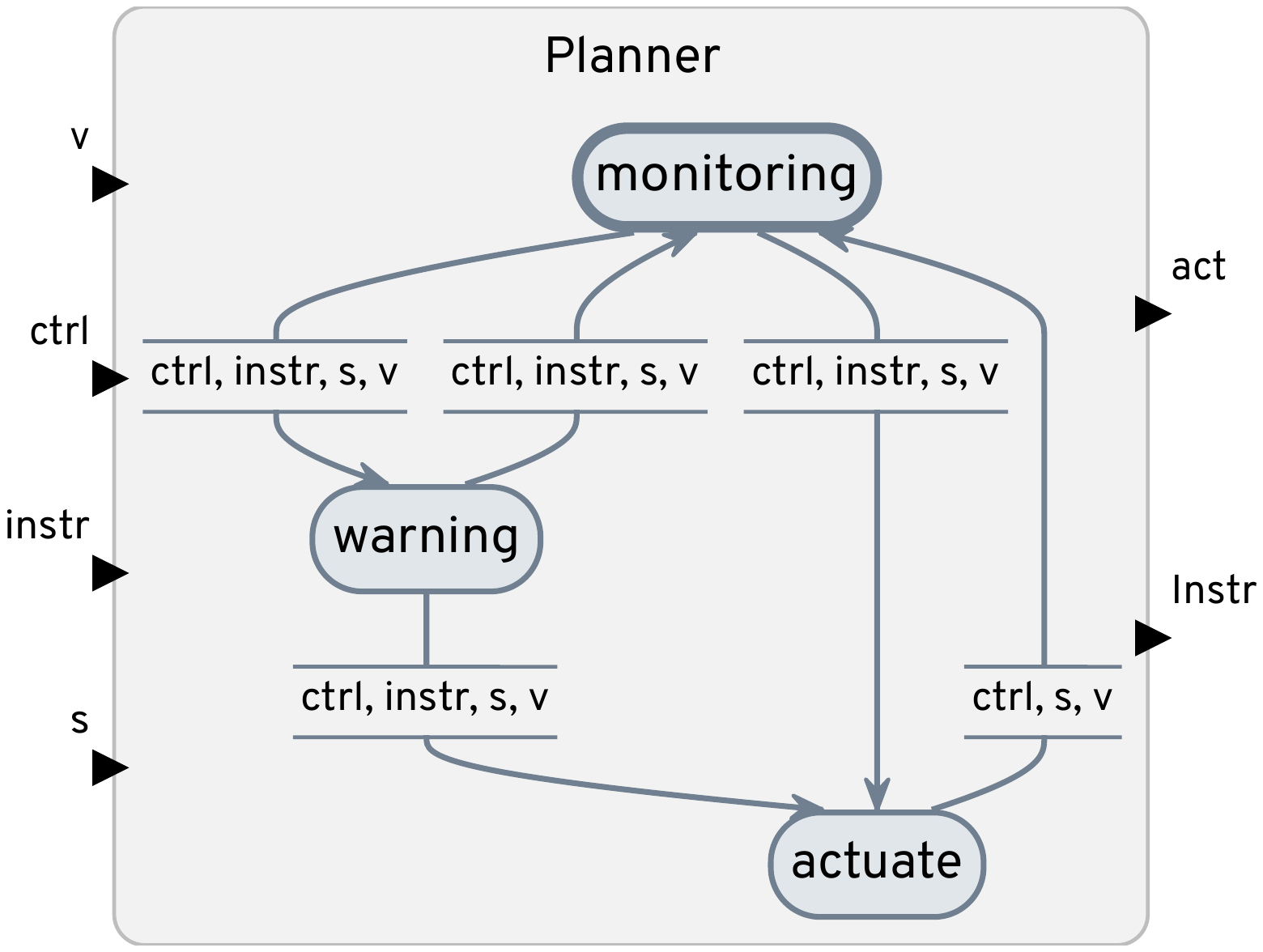}
\vspace{-10pt}
    \caption{The \texttt{Planner} reactor as a modal model where transitions are triggered by control signals defined in \tablename~\ref{tab:ctrl_tokens}.}
    \label{fig:agentreactor}
    
\end{figure}

\begin{figure}
    \centering
    \includegraphics[width=1\columnwidth]{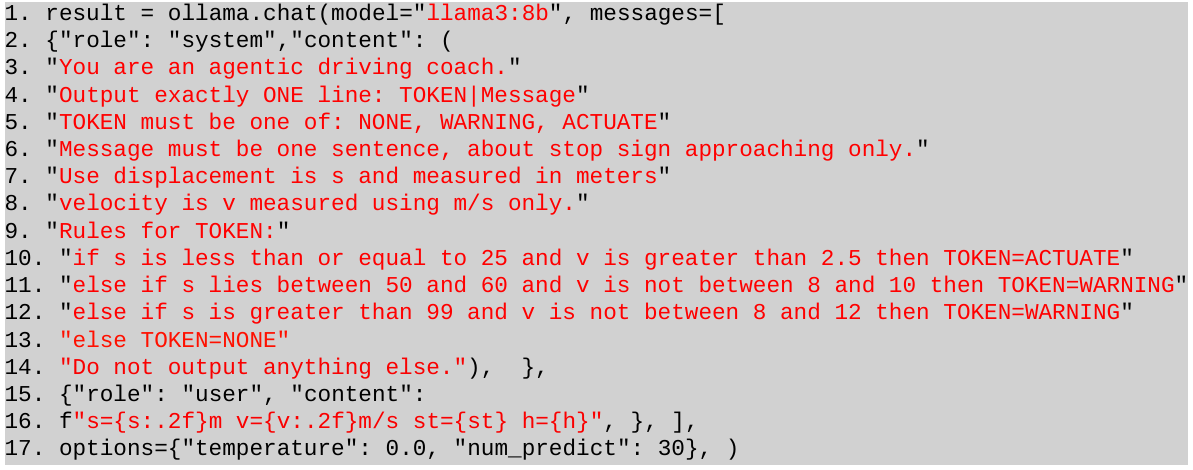}
\vspace{-10pt}
    \caption{An example structured prompt provided for \texttt{LLMInference} to generate a control signal (denoted as TOKEN) and a driving instruction for the \textbf{stop sign} scenario.}
    \label{fig:prompt}
    \vspace{-10pt}
\end{figure}

\textbf{\texttt{Planner}} (\figurename~\ref{fig:agentreactor}) implements an \textbf{LF modal model}~\cite{schulz2023polyglot} with three modes,  \texttt{monitoring}, \texttt{warning}, and \texttt{actuation}.
\textbf{\texttt{Planner}} takes displacement and velocity from the physical world as inputs, as well as the control and the instruction from \textbf{\texttt{LLMInference}}, then produces a processed instruction for the driver, \texttt{planned\_instr} as an output.
The \texttt{monitoring} mode, a default mode that can be entered by the \textit{NONE} signal, is active when the driver behavior remains within predefined safety bounds (e.g., within the safe velocity).
If the driver behavior deviates from the safety bounds but remains within a recoverable margin, the LLM produces \textit{WARNING} as a \texttt{ctrl} signal to cause \textbf{\texttt{Planner}} to enter the \textit{warning} mode and issue a corrective instruction.
We throttle the rate of corrective instructions to at most 1 instruction per second to prevent the driver from being overwhelmed by too frequent instructions~\cite{huang2024chatbot}. 
When the driver's behavior exceeds the recoverable margins, \textbf{\texttt{Planner}} enters the \textit{actuation} mode triggered by the \texttt{ctrl} signal \textit{ACTUATE}, and sends the actuation command via an output port, \texttt{act}, which is connected to the \texttt{Car} reactor.
This is similar to a human driving coach applying an emergency brake in the passenger seat.
We model the coach-to-actuation delay as a logical delay of 200ms.

\subsection{Structured Agentic Coach Prompt}
\label{sec:prompt}

In addition to our reactor model, we design a structured LLM prompt for the agentic coach to generate the control signals and driving instructions.
We design the prompt such that it reduces verbosity, ambiguity, and the risk of unsafe instructions by embedding tailored rules constraining responses from the agentic coach.


\figurename~\ref{fig:prompt} illustrates an example of our structured prompts for the stop sign scenario (one of our evaluation scenarios explained in Section~\ref{sec:scenarios}).
Our prompts follow a template provided by Ollama~\cite{marcondes2025using}. 
Our prompts consist of two components: a \textbf{system message} (Lines 3--14 in \figurename~\ref{fig:prompt}) that defines the behavioral rules for the agentic driving coach, and a \textbf{user message} (Line 16 in \figurename~\ref{fig:prompt}).

The \textbf{system message} specifies the required output format and safety rules. 
We instruct the LLMs to produce exactly one line in the format \texttt{Signal|Message} (Line 4 in \figurename~\ref{fig:prompt}). 
The accompanying message must be a single sentence describing the appropriate driving instruction (Line 6 in \figurename~\ref{fig:prompt}).
The \textbf{user message} (Line 16 in \figurename~\ref{fig:prompt}) provides the current driving context with the numerical and enumeration (Python \texttt{enum}) values of velocity, displacement, steer position, and head position, which are obtained from the environment reactor and the car reactor.
We enable context-aware reasoning of the LLMs by inserting runtime values generated by the \textbf{\texttt{Driver}} and \textbf{\texttt{Environment}} reactors during the execution of our reactor model program written in LF.
Also, we configure the LLMs to generate a maximum of 30 tokens (\texttt{"num\_predict": 30} at Line 17 of \figurename~\ref{fig:prompt}), while controlling randomness (\texttt{"temperature": 0}). 

\section{Evaluation}
We evaluate our agentic driving coach across a range of scenarios and LLMs, demonstrating deterministic, repeatable behavior. Our implementation and results are available in our GitHub repository.\footnote{\url{https://github.com/asu-kim/agentic-driving-coach.git}, which includes LF scenarios: \texttt{{StopSign.lf}}, \texttt{{SpeedChange.lf}}, and \texttt{LaneChange.lf}.}

\begin{figure}
    \centering
    \includegraphics[width=0.8\columnwidth]{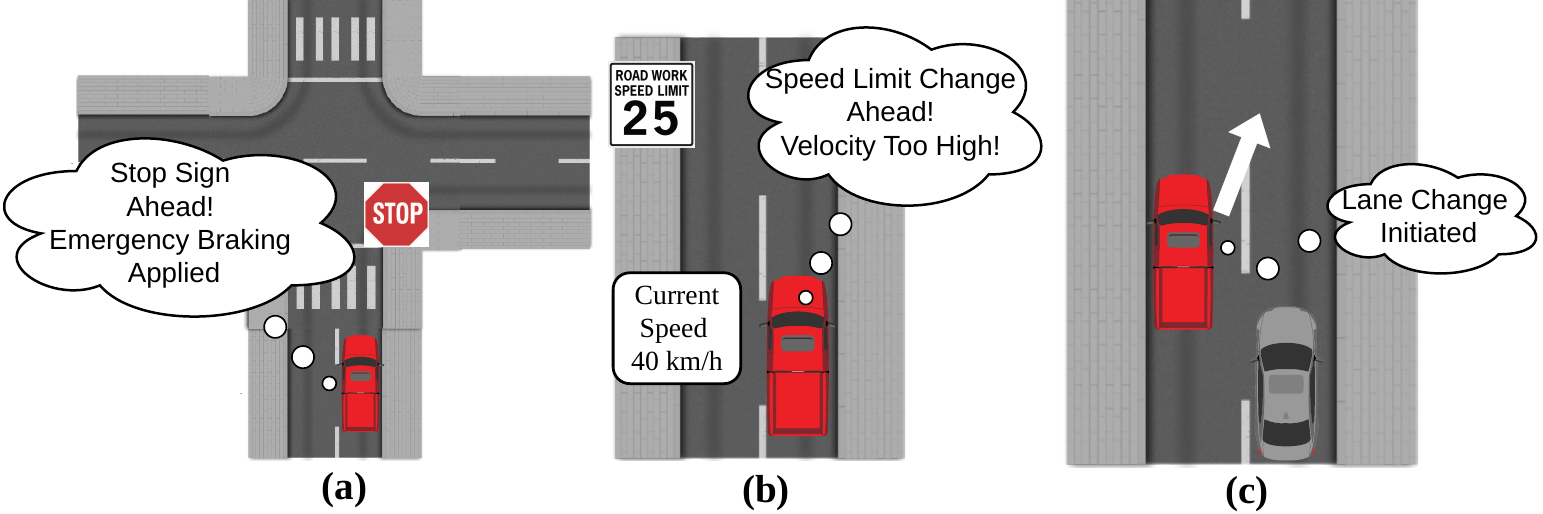}
    \caption{Three scenarios to evaluate the implementation of the proposed approach. (a) Stop sign, (b) Speed changing, and (c) Lane changing.}
    \vspace{-10pt}
    \label{fig:scenario}
\end{figure}

\begin{figure*}
    \centering
    \includegraphics[width=0.75\textwidth]{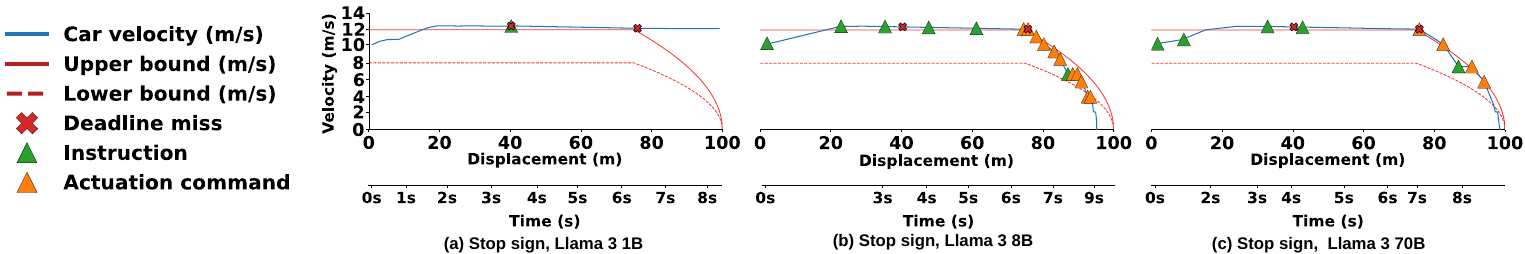}
    \vspace{-5pt}
   \caption{
   Evaluation results illustrating velocity vs. displacement and time for three Llama 3 models (1B, 8B, 70B) in a stop sign scenario. Blue lines show car velocity; solid/dashed red lines show safety bounds of car velocity. Red X marks deadline violations in \textbf{\texttt{LLMInference}}. Green and orange triangles indicate coach instructions and actuation commands, respectively.
   }
   \vspace{-10pt}
    \label{fig:stopbeginner}
\end{figure*}

\begin{figure*}
\centering
 \includegraphics[width=0.9\linewidth]{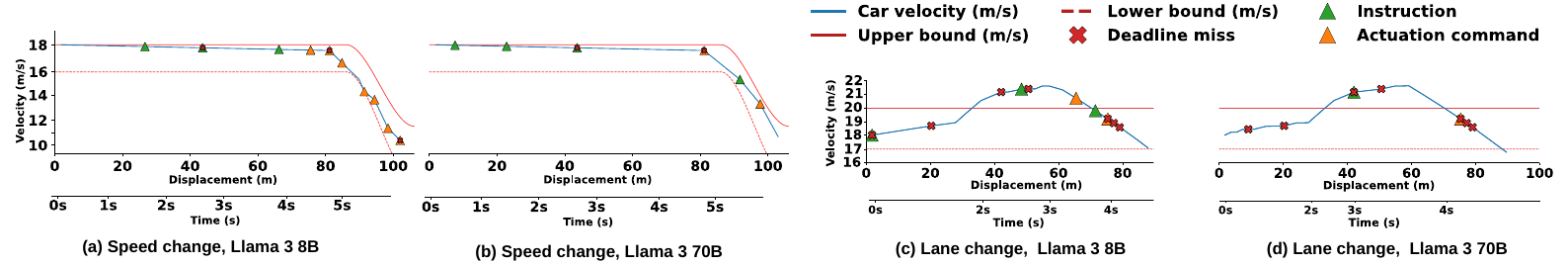}
    \vspace{-5pt}
\caption{
Evaluation results with Llama 3 (8B vs. 70B) in speed and lane change scenarios, illustrated by velocity vs. displacement and time.
}
    \vspace{-15pt}
\label{fig:stopinter}
\end{figure*}

\subsection{Evaluation Scenarios}
\label{sec:scenarios}
We use three driving scenarios shown in \figurename~\ref{fig:scenario}; (a) stop sign, (b) speed change, and (c) lane change, along with
a fixed, manually specified driver's behavior for each scenario.


\begin{compactItemize}
    \item \textbf{Stop sign}: A stop sign is located 100$m$ ahead of the car’s initial position.
    The car starts with an initial velocity of 10$m/s$ and must reach 0$m/s$ at the stop sign.
    We first define the desirable velocity such that it follows a half-parabolic trajectory to reach 0$m/s$ at the stop sign, and define the safe velocity bounds (upper and lower bounds) as $\pm$2$m/s$ of the desirable velocity.
    \item \textbf{Speed change}: Approaching a speed limit sign at 100$m$ distance, the car needs to decelerate from 18$m/s$ to 11$m/s$.
    \item \textbf{Lane change}: The car needs to change lanes to the right within 100$m$ while maintaining a velocity of 18$m/s$ to prepare for the right turn.
    The driver must check right, maintain speed, and steer into the right lane.
\end{compactItemize}

\subsection{Experimental Setup and Choice of Deadlines}
\label{sec:deadline_selection}

This section describes the experimental setup of our evaluation and how we measure the LLM inference latency and configure deadlines of the deadline handler of the \textbf{\texttt{LLMInference}} reactor in \figurename~\ref{fig:reactormodel}.

We use locally deployed 4-bit quantized Llama 3 models~\cite{grattafiori2024llama} with different numbers of parameters (1B, 8B, and 70B) using the Ollama runtime~\cite{marcondes2025using} running on a workstation with Intel Xeon w9-3475X (36 cores, 2.2 GHz to 4.8 GHz), 256 GB DDR5 main memory, and NVIDIA RTX 6000 Blackwell Pro GPU with a preloaded LLM.

We run the LLM inference of each model using our structured prompt tailored for each scenario 300 times, and choose the worst-case measured latency as the deadline of \textbf{\texttt{LLMInference}}'s deadline handler for the LF programs using the corresponding Llama model.
The measured worst-case latencies for the Llama 3 models with different numbers of parameters are: (a) 1B parameters: 186ms,  (b) 8B parameters: 250ms, and (c) 70B parameters: 613ms.

\subsection{Experimental Results and Discussion}

In this section, we discuss the evaluation results and their deterministic behaviors with the three Llama 3 models with 1B, 8B, and 70B parameters.
Overall, we confirm that for the same timing of inference and for the same control signals and driving instructions from the driving coach, as well as the same human driver's behavior, our reactor model always produces the same output behavior, including the deadline misses and handling, as well as the car's driving behavior in terms of the velocity and the displacement.
Thus, we empirically validate that our reactor model demonstrates the deterministic behavior for the same initial states (initial displacement and initial velocity) and the same series of inputs (timing and accuracy of the LLM inferences and human driver behavior).

While the overall car's driving behavior and the deadline misses in our reactor model vary with different inputs, we present representative experimental results in Figs.~\ref{fig:stopbeginner} and~\ref{fig:stopinter}.
Although we use the worst-case measured latency as deadlines, the results show some deadline misses, mainly because (a) the overhead and interference from LF program execution, (b) the LLM inference is frequently triggered every 100ms, and (c) the prolonged inference time when the driver's behavior and the car's velocity and acceleration change dynamically.
However, considering that the LLM inference is triggered every 100ms, the deadline misses are relatively rare.

\figurename~\ref{fig:stopbeginner} illustrates evaluation results for the stop sign scenario with three different-sized LLM models.
\figurename~\ref{fig:stopbeginner}a shows that the car fails to slow down and stop at the stop sign (at 100 m), because the Llama 3 1B fails to produce correct driving instructions or actuation commands.
We conclude that the inference accuracy of Llama 3 1B is dangerously low and unsuitable as the foundation model for an agentic driving coach, even though its inference latency is lower than that of larger Llama models.
\figurename~\ref{fig:stopbeginner}b and \figurename~\ref{fig:stopbeginner}c show the results using the Llama 3 8B and 70B models, respectively.
We find that the larger model (70B) produces fewer driving instructions and actuation commands, while letting the driver stop the car nearer to the stop sign compared to the smaller model (8B), indicating that the larger, more accurate foundation model leads to safer behavior in agentic HITL CPS.

\figurename~\ref{fig:stopinter} shows the evaluation results for the speed change and lane change scenarios, using Llama 3 8B and 70B models.
We exclude the 1B model results from the figure, as it also leads to unsafe driving behavior in these two scenarios.
As shown in the results from the speed change scenario in \figurename~\ref{fig:stopinter}a and \figurename~\ref{fig:stopinter}b, again the larger model (70B) successfully guides the student driver to reduce the speed to the new speed limit with fewer instructions and actuation commands, although both models encounter deadline misses, which trigger fallback mechanisms in the deadline handler to be engaged.
\figurename~\ref{fig:stopinter}c and \figurename~\ref{fig:stopinter}d illustrate the experimental results for the lane change scenario, we find that there are more deadline misses compared to the other two scenarios, partially due to the increased overhead of the coach to monitor the head position/orientation of the driver (to check if the driver checks the right lane before steering) as well as the car's velocity and acceleration.
Similar to the other two scenarios, the larger model (70B) leads the human driver into the right lane with fewer instructions and actuation commands than the smaller model (8B).

\section{Conclusion}
In this paper, we explore a methodology to ensure the robustness and determinism of agentic AI-powered HITL CPS, using a concrete use case: an agentic driving coach modeled with the reactor MoC and the Lingua Franca (LF) framework.
Our reactor MoC-based agentic driving coach model achieves deterministic behavior in response to the same inputs of the driver's behavior and the coach's control signals and driving instructions, in terms of timing and accuracy.
Our reactor model provides a timely and safe fallback mechanism in case the LLM inference is delivered late, with hard-coded rules.
As future work, we will build an integrated, single LF program for an agentic driving coach that supports multiple scenarios.
Furthermore, we will incorporate advanced LLM fine-tuning techniques such as retrieval-augmented generation (RAG)~\cite{prahlad2025personalizing} or low-rank adaptation (LoRA)~\cite{zhao2024galore} to reduce inference latency while providing LLM agents with context-enriched prompts.
We will also integrate our hardware setup and evaluate our agentic driving coach with real-time sensor inputs from human drivers to further study common human behaviors across a wider range of scenarios. 

\section*{Acknowledgment}

This work was supported in part by NSF under Award No. POSE-2449200 (An Open-Source Ecosystem to Coordinate Integration of Cyber-Physical Systems).

\bibliographystyle{IEEEtran}
\bibliography{refs}

\end{document}